\documentclass[preprint,12pt,authoryear]{elsarticle}

\usepackage{amssymb}
\usepackage{amsmath}
\usepackage[english]{babel}
\usepackage{csquotes}
\usepackage{float}
\usepackage{graphicx} 
\usepackage{chngcntr}
\counterwithout{figure}{section}

\journal{Engineering Applications of Artificial Intelligence}
\usepackage[printonlyused]{acronym}

\acrodef{cbm}[CBM]{Condition-Based Maintenance}
\acrodef{dl}[DL]{Deep Learning}
\acrodef{ml}[ML]{Machine Learning}
\acrodef{cnn}[CNN]{Convolutional Neural Network}
\acrodef{nn}[NN]{Neural Network}
\acrodef{ann}[ANN]{Artificial Neural Networks}
\acrodef{lstm}[LSTM]{Long-Short Term Memory}
\acrodef{rnn}[RNN]{Recurrent Neural Network}
\acrodef{auc}[AUROC]{Area Under the Curve Receiver Operating Characteristics}
\acrodef{lstm}[LSTM]{Long-Short Term Memory}
\acrodef{ai}[AI]{Artificial Intelligence}
\acrodef{nlp}[NLP]{Natural Languaje Processing}
\acrodef{cbtc}[CBTC]{Communication-Based Train Control}
\acrodef{wa}[WA]{Wheel Assembly}
\acrodef{mems}[MEMS]{Micro-Electromechanical Systems}
\acrodef{fbg}[FBG]{Fiber Bragg Grating}
\acrodef{aba}[ABA]{Axle Box Acceleration}
\acrodef{svr}[SVR]{Support Vector Regression}
\acrodef{gan}[GAN]{Generative Adversarial Network}
\acrodef{vae}[VAE]{Variational Autoencoder}
\acrodef{gru}[GRU]{Gated Recurrent Unit}
\acrodef{relu}[ReLU]{Rectified Linear Unit}
\acrodef{bptt}[BPTT]{Backpropagation-Through-Time}
\acrodef{elt}[ELT]{Enhancion of the Locality of Transformer}
\acrodef{lse}[LSE]{Long-Sum-Exp}
\acrodef{mse}[MSE]{Mean Squared Error}
\acrodef{mae}[MAE]{Mean Absolute Error}
\acrodef{mape}[MAPE]{Mean Absolute Percentage Error}
\acrodef{sgd}[SGD]{Stochastic Gradient Descent}
\acrodef{lr}[LR]{Learning Rate}
\acrodef{ecg}[ECG]{Electrocardiography}
\acrodef{mlp}[MLP]{Multilayer Perceptron}
\acrodef{ndvi}[NDVI]{Normalized Difference Vegetation Index}
\acrodef{stft}[STFT]{Short Time Fourier Transform}
\acrodef{istft}[iSTFT]{Inverse Short Time Fourier Transform}
\acrodef{ssf}[SSF]{Spectral ShaftFormer}
\acrodef{sf}[SF]{ShaftFormer}
\acrodef{tpe}[TPE]{Tree-structured Parzen Estimator}
\acrodef{lhs}[LHS]{Left Hand Size}
\acrodef{rhs}[RHS]{Right Hand Size}
\acrodef{LOESS}[LOESS]{Locally Estimated Scatterplot Smoothing}

\begin{document}

\begin{frontmatter}

\title{Transformer Vibration Forecasting for Advancing Rail Safety and Maintenance 4.0}

\author{Darío C. Larese\corref{cor1}\fnref{label1}}

\author{Almudena Bravo Cerrada\corref{cor1}\fnref{label1}}

\author{Gabriel Dambrosio Tomei\fnref{label2}}
\ead{gdambros@pa.uc3m.es}

\author{Alejandro Guerrero-López\fnref{label3}}
\ead{alejandro.guerrero@upm.es}

\author{Pablo M. Olmos\fnref{label1}}
\ead{pamartin@ing.uc3m.es}

\author{María Jesús Gómez García\fnref{label2}}
\ead{mjggarci@ing.uc3m.es}

\cortext[cor1]{Both authors contributed equally to this work}

\address[label1]{Department of Signal Processing and Communications, Universidad Carlos III de Madrid, Leganés, 28911, Spain}
\address[label3]{Escuela Técnica Superior de Ingenieros de Telecomunicación, Universidad Politécnica de Madrid, 28040, Madrid, Spain}
\address[label2]{Department of Mechanical Engineering, Universidad Carlos III de Madrid, Leganes, 28911, Spain}


\begin{abstract}



Maintaining railway axles is critical to preventing severe accidents and financial losses. The railway industry is increasingly interested in advanced condition monitoring techniques to enhance safety and efficiency, moving beyond traditional periodic inspections toward Maintenance 4.0.

This study introduces a robust Deep Autoregressive solution that integrates seamlessly with existing systems to avert mechanical failures. Our approach simulates and predicts vibration signals under various conditions and fault scenarios, improving dataset robustness for more effective detection systems. These systems can alert maintenance needs, preventing accidents preemptively. We use experimental vibration signals from accelerometers on train axles.

Our primary contributions include a transformer model, \ac{sf}, designed for processing time series data, and an alternative model incorporating spectral methods and enhanced observation models. Simulating vibration signals under diverse conditions mitigates the high cost of obtaining experimental signals for all scenarios. Given the non-stationary nature of railway vibration signals, influenced by speed and load changes, our models address these complexities, offering a powerful tool for predictive maintenance in the rail industry.

\end{abstract}


\end{frontmatter}

\section{Introduction}

Train operations rely heavily on the reliability of their wheelsets, with axles being crucial for supporting train weight and transmitting power. Malfunctions can lead to severe consequences such as accidents, derailments, and safety risks, often due to fatigue or cracks in the axles, which can result in loss of life and property damage \citep{survey}.

To address these challenges, we propose a novel transformer model for predictive maintenance that forecasts a set of axle vibration signals obtained experimentally under different conditions of load, speed and flaws. In the field of predictive maintenance, forecasting is a crucial requirement for detecting faulty signals and simulating new signals which cannot be obtained easily. Through monitoring relevant variables, it is a natural solution to axle reliability issues.

Previous research has explored using various sensors to detect wheel flaws \citep{alemi} and predict failures in critical rail car components \citep{li2014improving, 9721893}. For railway axles, studies have utilized model-based signals from analytical \citep{gomezMMT1} or numerical models \citep{polimi1}, as well as experimental signals from scaled rigs \citep{gomezMMT2} or accelerometers on real systems. Despite the potential of these methods, bogie test rigs are rare and costly, limiting the scope and generalizing capabilities of the results \citep{gomezIJAV, gomezMMT1, gomezMMT2, polimi2, Gomez2023}.

The current trend in maintenance automation, known as Maintenance 4.0, leverages Machine Learning (ML) techniques. Recently, there has been growing interest in applying \ac{dl} for condition monitoring and predictive maintenance in the rail industry. Studies have demonstrated the use of neural networks to analyze vibration signals and predict axle failures \citep{GomezRESS, Gomez2018, Gomez2020}. For example, \citet{galdo2023detecting} explored classifying crack conditions on railway axles using a differential convolutional classification network. However, existing studies often overlook the nonstationary nature of railway vibration signals, which are influenced by speed, load, and environmental changes \citep{YE2022111268, 8006280}. This oversight hampers the development of models that can generalize across diverse operating conditions and structural variations.

Moreover, many current methods rely on traditional ML algorithms or simplistic neural network architectures that may struggle to capture the complex relationships present in railway vibration data \citep{Serin2020}. Some research has proposed \ac{dl} models for machine diagnosis, but these often lack domain-specific adaptations tailored to railway systems \citep{ZHAO2019213, Wang2021}. Simple transformer models have been applied to wheel defect classification tasks \citep{9721893}, yet they remain plug-in solutions rather than tailored DL models.



This study aims to address these gaps by proposing the \ac{sf} model and the \ac{ssf} model, which are variants of the original Transformer \citep{vaswani2017attention}. These models implicitly consider the non-stationarity of the data and effectively model the signals. By incorporating domain knowledge from railway engineering, our methodology is designed to fit within the monitoring and maintenance tasks in the industry. Moreover, both models are able to generate signals under different environmental conditions such as load and speed. Potential applications of this study include outlier detection, missing data imputation, signal forecasting, and data augmentation.

\section{Methodology}
\subsection{Data collection: experimental setup}

\begin{figure}[t] \centering
    \includegraphics[width=0.75\columnwidth]{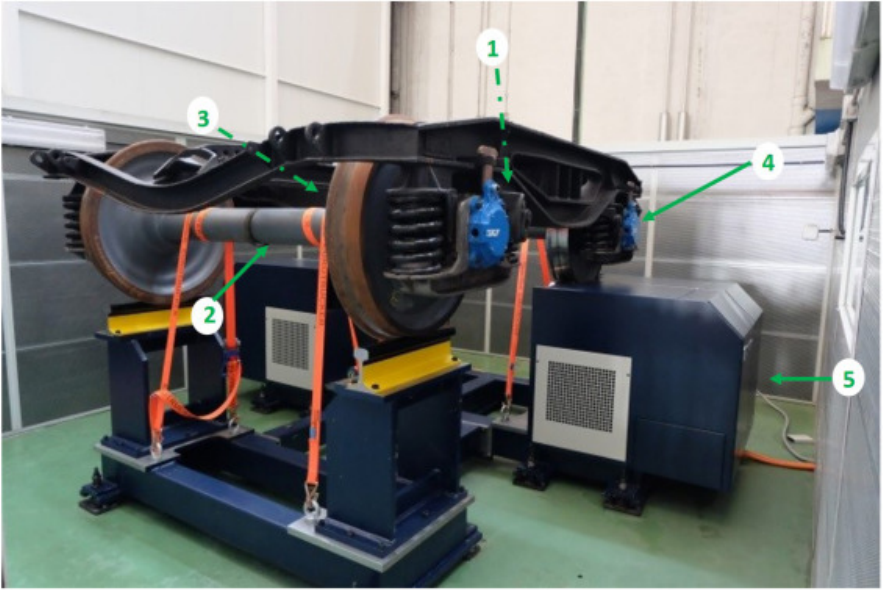}
    \caption[bogie Front View]{Test rig with Bogie Y21 set, including tested wheelset (1), fixed wheelset (2), \ac{lhs} axle box (3), \ac{rhs} axle box (4), motor and driving system (5)}
    \label{fig:bogie1}
\end{figure}

The experimental acquisition of vibration signals from railway axles involved investigating the dynamics of cracked rotors using a specially designed experimental setup, as detailed in previous studies \citep{Gomez2018, Gomez2020}. The experimental system was constructed within a bogie test rig, consisting of a fixed bench and a drive system to roll the tested axle, as shown in Figure \ref{fig:bogie1}.

Six accelerometers were placed, with three in each axle box (\ac{lhs} and \ac{rhs}) of the tested axle, measuring vibrations in three orthogonal directions: tangential (in two perpendicular directions), vertical (aligned with the gravitational force) and axial (parallel to the direction of the track). In addition, the rig featured a loading system that employs hydraulic actuators to apply a consistent vertical load to the bogie via a chain mechanism. Throughout testing, the selected load remained constant to accurately replicate real-world operating conditions.

The experiments involved testing three distinct Wheelset Assemblies (WAs), labeled WA1, WA2 and WA3, resulting in the creation of three individual datasets. Each WA was subjected to testing under two different combinations of speed and load, for clockwise and counterclockwise rotations. The applied load and speed were kept constant throughout the tests and are outlined in \ref{table:crackdata}, offering a comprehensive overview of all test conditions.

Monitoring these signals on a bogie instead of a wheel set produces very rich data, as the full bogie includes not only the wheels and axle, but also the suspension systems and load, reproducing with high fidelity the reality.

\begin{table}[t] 
    \centering
    \begin{tabular}{|l|l|}
        \hline
        \textbf{Test condition} & \textbf{Values} \\
        \hline
        Flaw condition & Healthy shaft (D0) \\
                       & 5.7 mm (D1) \\
                       & 10.9 mm (D2) \\
                       & 15 mm (D3) \\
        \hline
        Load & 4t \\
             & 10t \\
        \hline
        Speed & 20 km/h \\
              & 50 km/h \\
        \hline
        Wheelsets rotation & Counterclockwise \\
                           & Clockwise \\
        \hline
        Axle box accelerometer orientation & Lengthwise \\
                                           & Vertical \\
        \hline
        Axle box accelerometer place & \ac{rhs} \\
                                     & \ac{lhs} \\
        \hline
    \end{tabular}
    \caption{Conditions tested for each WA.}
    \label{table:confs}

\end{table}

The test rig comprises a fixed bench and a drive system for rolling the tested axle and also incorporates a loading system that uses hydraulic actuators that apply a vertical load to the bogie through a chain. During the test, the load remains constant, replicating in real-world situations. Table \ref{table:confs} shows the tested configurations.

First, each axle was tested under control conditions. Following this, without disassembling the axle, different sizes of flat-front transverse cracks were machined at the central position of the axle. A total of three different depths of crack were introduced, resulting in testing four different crack conditions. These conditions are measured in terms of depth (mm) and percentage of damaged depth relative to axle diameter (\%). The corresponding values are shown in Table \ref{table:crackdata}.

The axles tested are made of steel EA1N, with a total diameter of 170 mm, and a total length of 2.433 m.

To complement the mechanically obtained signals, this study incorporates numerical simulations of expected signal behavior using the Abaqus software. Once the Y21 shaft model was created in Abaqus with the specified design conditions, simulations were performed for various loads and speeds. To obtain vibration signals in the frequency domain, a specific point on the shaft was selected to collect displacement data throughout the simulation. For these simulations, all measurements were taken at the left node of the shaft, as highlighted in the accompanying figure. 

Using the numerical data, a frequency-domain analysis was conducted to compare the operating conditions of the shaft. This analysis was performed using the MATLAB numerical computing system. Cracks in the shaft theoretically induce frequency peaks corresponding to the shaft's rotational speed and its harmonics. Through this analysis, these changes can be observed, enabling quantitative measurements of the effects. The integration of simulated data with mechanical measurements provides a comprehensive approach to understanding and diagnosing the dynamic behavior of the shaft under varying conditions.

Cracks in the shaft theoretically cause peaks in the rotational speed of the shaft and its respective harmonics \citep{ElArem2019}. With this analysis, these changes can be observed and a quantitative measurement of them can be obtained. These numerical data can be used to further enrich the dataset from a theoretical and experimental point of view.

\begin{table}[t]
\centering
    {\begin{tabular}{lcc}
        \hline
        {\textbf{Crack Level}} & {\textbf{Depth (mm)}} & {\textbf{Damaged Depth (\%)}} \\
        \hline
        D0 & 0 & 0 \\
        D1 & 5.7 & 0.03 \\
        D2 & 10.9 & 0.06 \\
        D3 & 15 & 0.08 \\
        \hline        
    \end{tabular}}
    {\caption{Axle Crack Data. D0 corresponds to a healthy axle. D1, D2, and D3 correspond to the increasing depths of the cracks of the damaged axle} \label{table:crackdata}}

\end{table}

\subsection{The \ac{sf} Model}

\begin{figure}[t]
    \centering

    \includegraphics[width=\textwidth]{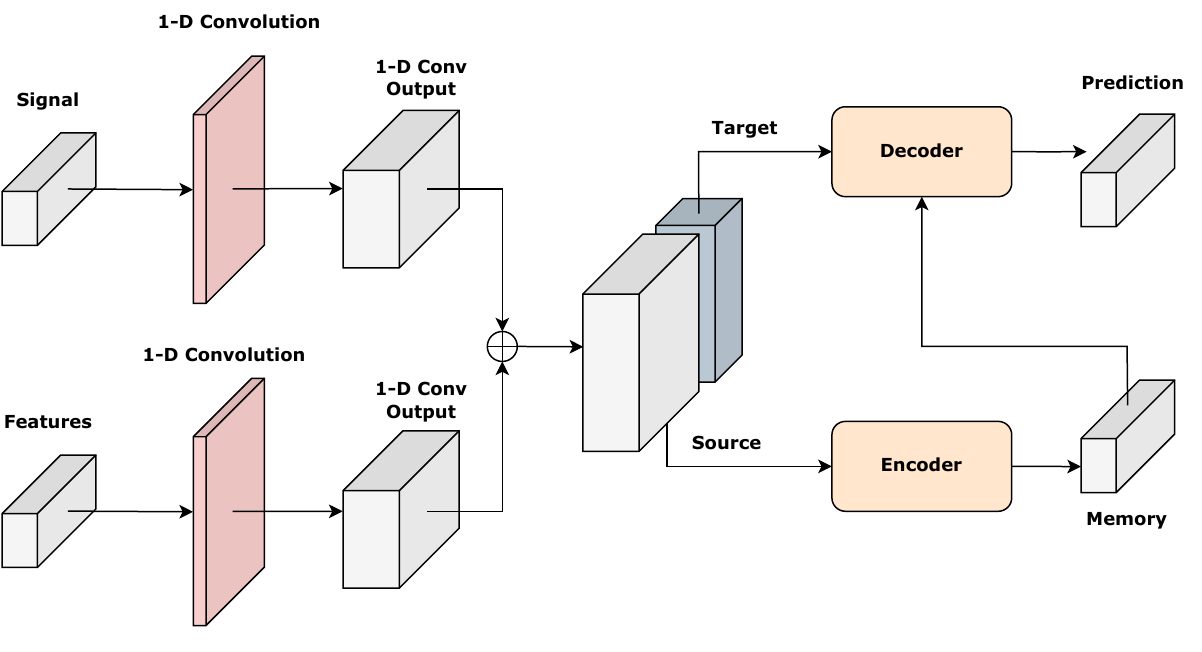}
    \caption[Architecture \ac{sf} for forecasting]{Architecture of the \ac{sf}, showing 1) \ac{elt}: the preprocessing layers for the signals and their features, 2) the encoder and the generated encoder memory, 3) the decoder and the resulting prediction. The input of the encoder, called \textquote{source}, corresponds to historical data, while the input of the decoder, the \textquote{target}, corresponds to the true values of the future time steps which are going to be predicted. This method is called teacher forcing and is used when training the model. It is used widely in transformer architectures \citep{vaswani2017attention}.}
    \label{fig:arch}

\end{figure}

We present the \ac{sf}, a model that extends the Informer \citep{zhou2021informer} model by means of an \acf{elt} preprocessing step. \ac{elt} identifies the relevant information from the 1-D input vibration signal by passing it through a set of \ac{cnn} and augmenting the dimension of the input signal.

The \ac{sf} incorporates the ProbSparse attention mechanism introduced by the Informer, a method used to address the challenges of processing long sequences efficiently. It reduces computational complexity by sparsifying the attention matrix, selectively focusing on more relevant parts of the input data, rather than computing attention scores for every possible pair of inputs. This method helps to manage memory more effectively and improve computational efficiency, particularly suitable for models dealing with long input sequences, such as vibration signals, as it avoids the quadratic complexity typical of standard self-attention mechanisms. It creates a powerful encoder memory, ensuring that it contains the relevant data from the history of the signal. Figure \ref{fig:arch} shows the general architecture of this model.

\begin{figure}[t]
    \centering
    \includegraphics[width=\textwidth]{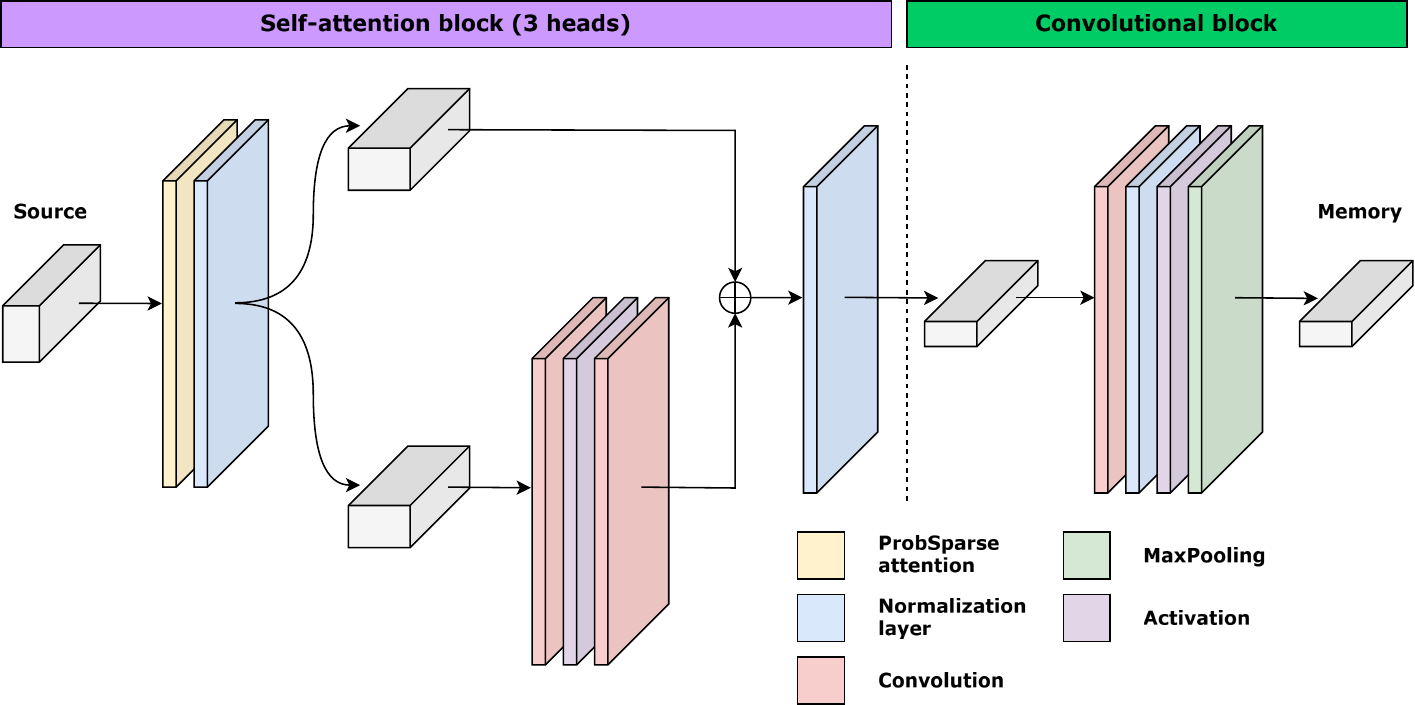}
    \caption[Architecture of the \ac{sf}'s encoder]{Architecture of the \ac{sf}'s encoder based on the Informer, using the ProbSparse self-attention layer, in purple, and the convolutional block, in green.}
    \label{fig:encoder}
\end{figure}

\paragraph{Encoder} The encoder contains two different blocks, shown in Figure \ref{fig:encoder}, which are the self-attention block and the convolution block. The first block is composed of the ProbSparse self-attention layers, from which we obtain the attention values. We make two copies of these values. One copy will go through a down-sample convolution layer and then an up-sample convolution layer with the intention of learning the information and then being able to reproduce it. Then, the two copies, the residual and the modified, are concatenated and processed by the convolutional block. After a convolutional layer, a pooling layer is applied to down-sample the data obtaining the memory. The memory maintains the general information of the signals and is passed to the decoder. 

\paragraph{Decoder} We use a vanilla Transformer decoder \citep{vaswani2017attention} to process this memory. The \ac{sf} is posed as an autorregressive generative model. The input to this model is the signal in the time domain. For each time step in the future, the model generates a probability distribution over the next value of the signal. The loss function used is the \ac{mse}. We use a Normal distribution as the output distribution, parameterized by its mean and variance, and allow for gradient backpropagation using the reparameterization trick \citep{reparam}. The mathematical details can be found at \ref{appendix:reparam_trick}. We also add white Gaussian noise to estimate a potential interference in the signal reconstruction, inspired by \citet{white1} for computer vision applications. Finally, this distribution is sampled, yielding the predicted signal.

\begin{figure}[H]
    \centering
    \includegraphics[width=6cm]{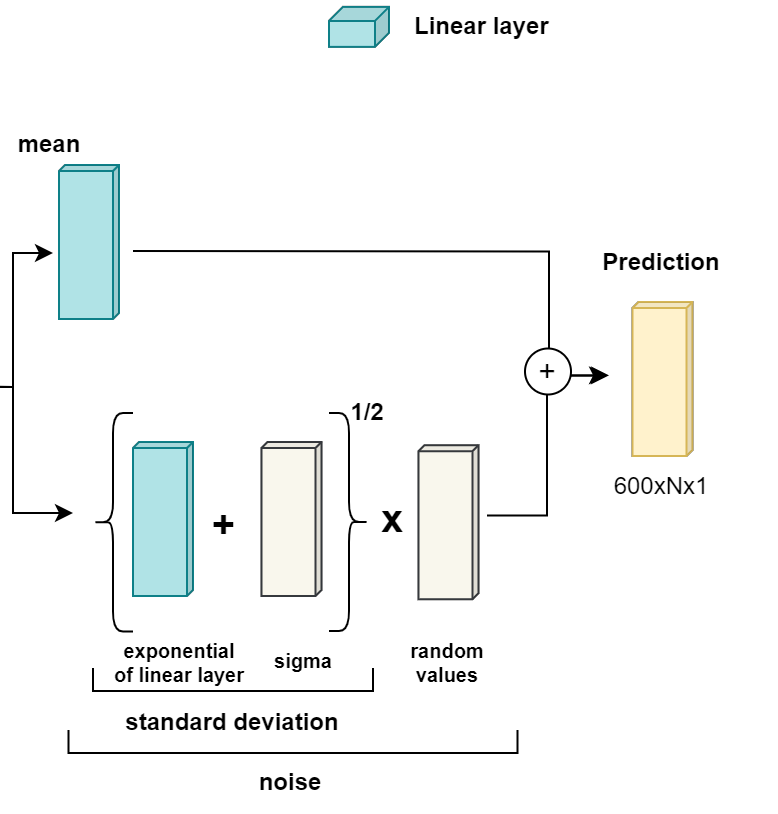}
    \caption[Output of the \ac{sf} model]{Output of the \ac{sf} model}
    \label{fig:only_noise}
\end{figure}

Using a probabilistic model, we not only obtain an estimated central prediction but also measure and quantify the uncertainty associated with that prediction.

The implementation of the \ac{sf} model can be found on GitHub\footnote{https://github.com/AlmudenaBravoC/ShaftFormer}.

\subsection{The \ac{ssf}}

In time series forecasting, working in the frequency domain offers valuable advantages by providing insights into the periodic patterns, noise filtering, and spectral analysis of the data. The transformation of time-series data into frequency domain is achieved through the \ac{stft}. It allows the identification of dominant frequencies and helps in the extraction of relevant features, such as amplitude and phase information. Additionally, the frequency domain allows for the explicit modeling of periodic components that may be challenging to capture in the time domain alone.

This model adopts the \ac{sf} model as a base to build upon. We do the following changes over the \ac{sf}:

\begin{itemize}
  \item Work in the frequency domain and select the most important frequencies using global filtering \citep{moreno2023deep}.
  
  \item Adapt the transformer encoder and the decoder using HiLo attention \citep{NEURIPS2022_5d5f703e}.
  
  \item Incorporate system to condition the model to the theoretical model, using sample signals generated using the finite element method.
  
  \item Modify the sampling model. The samples still come from a Normal distribution, but the variance of this distribution is now sampled from an exponential model to regularize too high variances.
  
\end{itemize}

The input signal is transformed to its spectrogram using a \ac{stft} of a certain time and frequency resolution. These resolutions are treated as hyperparameters and are optimized through cross-validation. Computing the \ac{stft} gives complex numbers as output. Both the real and imaginary part of the complex numbers are treated as independent channels, so the one-dimensional convolutional filters applied to each of them use a different kernel. 

\begin{figure}[H] \centering
    {\includegraphics[width=\columnwidth]{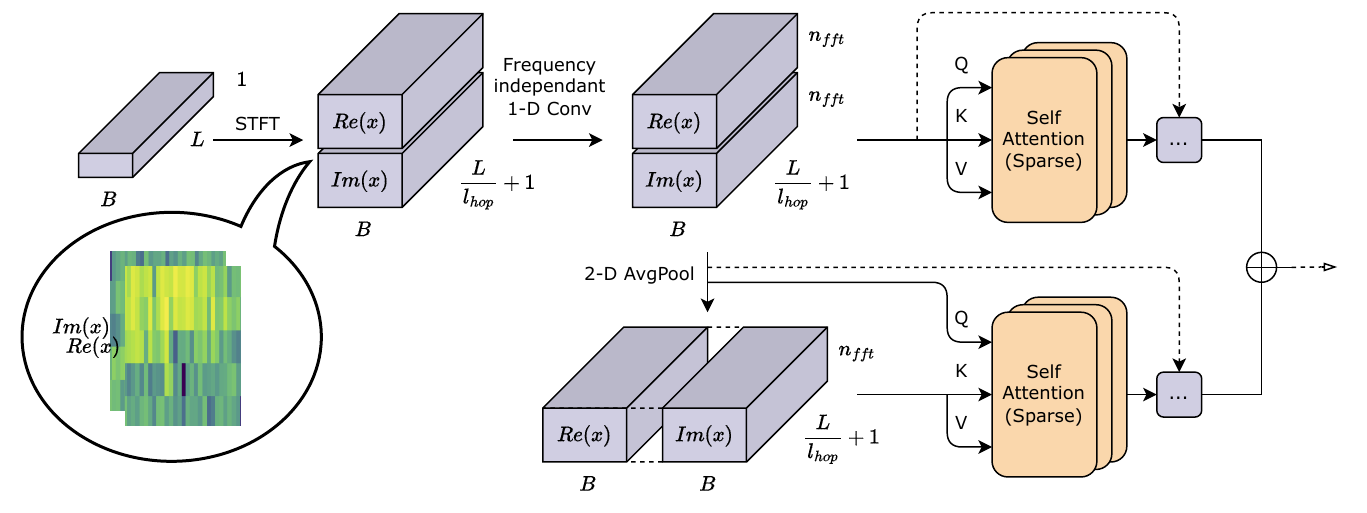}}
    \caption[Encoder of the \ac{ssf}]{Encoder of the \ac{ssf}}
    \label{fig:ssf_enc}
\end{figure}

\paragraph{Encoder} The encoder is modified by splitting the attention heads into two groups, as shown in Figure \ref{fig:ssf_enc}. The purpose of this modification is to efficiently capture higher and lower-level attention patterns and to reduce the impact of noise, as demonstrated by \citep{NEURIPS2022_5d5f703e}. The first group processes the spectrogram of the signal (high frequencies), while the second group processes a low-pass filtered version of the spectrogram (low frequencies). The low frequency attention block applies a 2-D Average Pooling to each of the components before applying the sparse attention mechanism. In contrast, the high-frequency attention block works with the full resolution of the spectrogram. As the last step in the encoder, the model concatenates the attention scores extracted of high and low frequencies.

The highly informative latent space which is generated using the encoder allows to use a simpler decoder.

\begin{figure}[H] \centering
    {\includegraphics[width=\columnwidth]{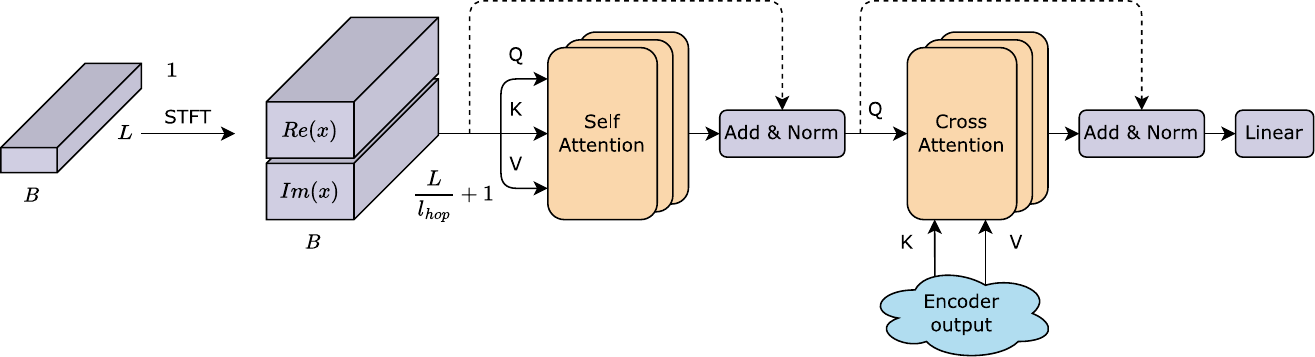}}
    \caption[\ac{ssf} decoder]{Decoder of the \ac{ssf}}
    \label{fig:ssf_dec}
\end{figure}

\begin{figure}[H] \centering
    {\includegraphics[width=0.65\columnwidth]{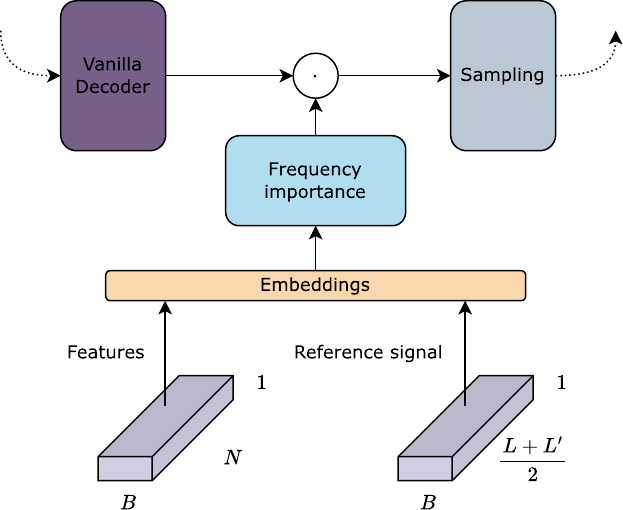}}
    \caption[Second part of the decoder in the \ac{ssf}]{Decoder in the \ac{ssf} (II): Feature inclusion, Frequency importance and Sampling}
    \label{fig:ssf_dec2}
\end{figure}

\paragraph{Decoder} The \ac{ssf} is posed as an autorregressive generative model. The input to this model is the signal spectrogram. For each time step in the future, the model generates a probability distribution over the next time window of the spectrogram. The loss function we use is the \ac{mse}. We choose a 2-step decoder where the first step is similar to the vanilla decoder from \citet{vaswani2017attention} as depicted in \ref{fig:ssf_dec}. The second part contains a frequency filtering mechanism, as shown in Figure \ref{fig:ssf_dec2}, which allows the model to selectively mute certain frequency bands, partially or completely, as proposed in \citet{moreno2023deep}. This approach enhances the model's ability to focus on the most relevant spectral features while reducing the influence of noise or irrelevant frequencies, leading to more accurate and context-aware predictions in the frequency domain. The decoder processes the reference signal and the features and merges them with the rest of the decoder. Additionally, it contains the sampling process and is responsible for providing the next values of the time series (the next time window in the spectrogram). At the very end, the signal is transformed back into the time domain using the \ac{istft}. Nevertheless, we evaluate the loss function in the frequency domain.

\paragraph{Sampling process}

To ensure that the sample variance is positive, to regularize too high variances, and due to the nature of this problem, we prefer to generate a signal which is close to the mean rather than generating outliers. For this purpose we use the following sampling scheme

\[
\begin{aligned}
    z & \sim \mathcal{N}(\mu(x), \sigma^2) \\
    \sigma^2 & \sim \text{Exp}(\lambda(x))
\end{aligned}
\]
where $x$ are the inputs and $z$ is the predicted value. Since stochastic sampling is not a differentiable process, the reparameterization trick is used together with the inverse transform sampling method to allow for proper gradient propagation \ref{fig:sampling}. See \ref{appendix:inverse_transform} for further details.

\begin{figure}[!t] \centering
    {\includegraphics{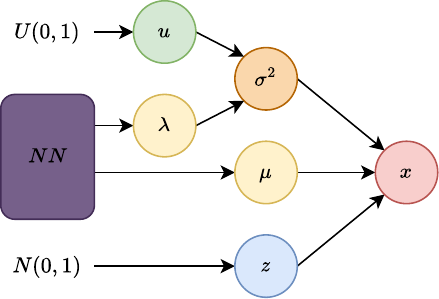}}
    \caption[Sampling]{Sampling process. The inverse transform sampling method is used to sample the variance given $\lambda$. The reparameterization trick generates the final sample using this variance and the predicted mean.}
    \label{fig:sampling}
\end{figure}

See \ref{appendix:training} for details on the training of this model.
The implementation of the \ac{ssf} model can be found on GitHub\footnote{https://github.com/dariocb/SpectralShaftFormer}.

\section{Results}
The results are divided into three main aspects: frequency domain analysis, time domain analysis, and signal decomposition.

\subsection{\ac{ssf} forecasting results}

\begin{figure}[H] \centering
    {\includegraphics[width=\columnwidth]{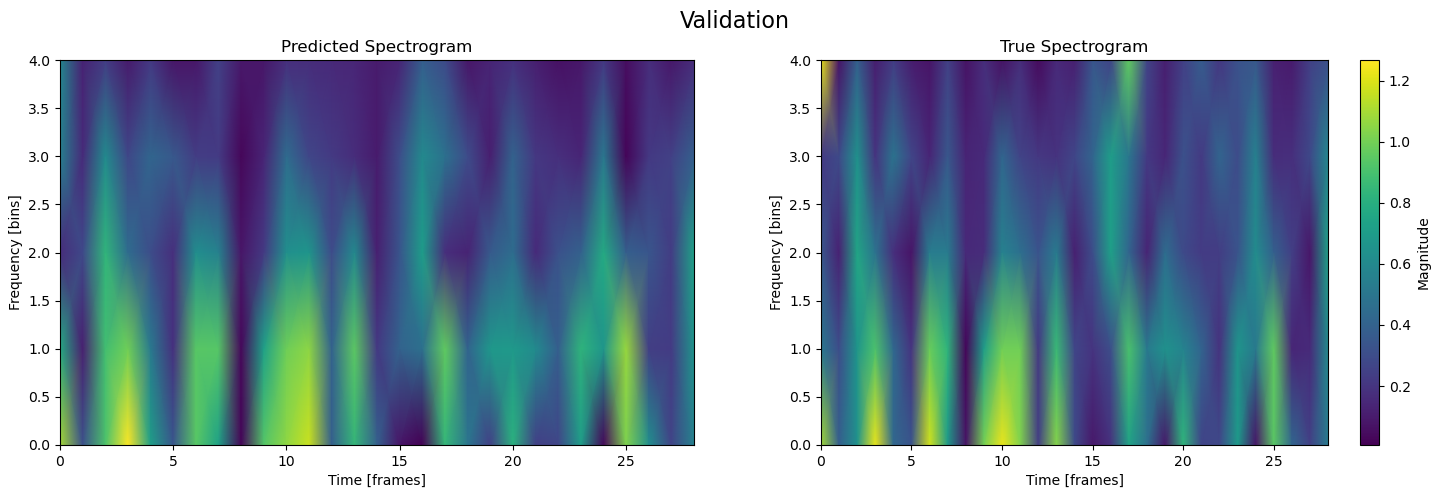}}
    \caption[Predicted spectrogram (validation)]{Predicted and true spectrogram. Validation results (\ac{mse}= 0.18). The horizontal axis represents time and the vertical axis represents frequency. }
    \label{fig:spec_valid}
\end{figure}

\begin{figure}[H] \centering
    {\includegraphics[width=\columnwidth]{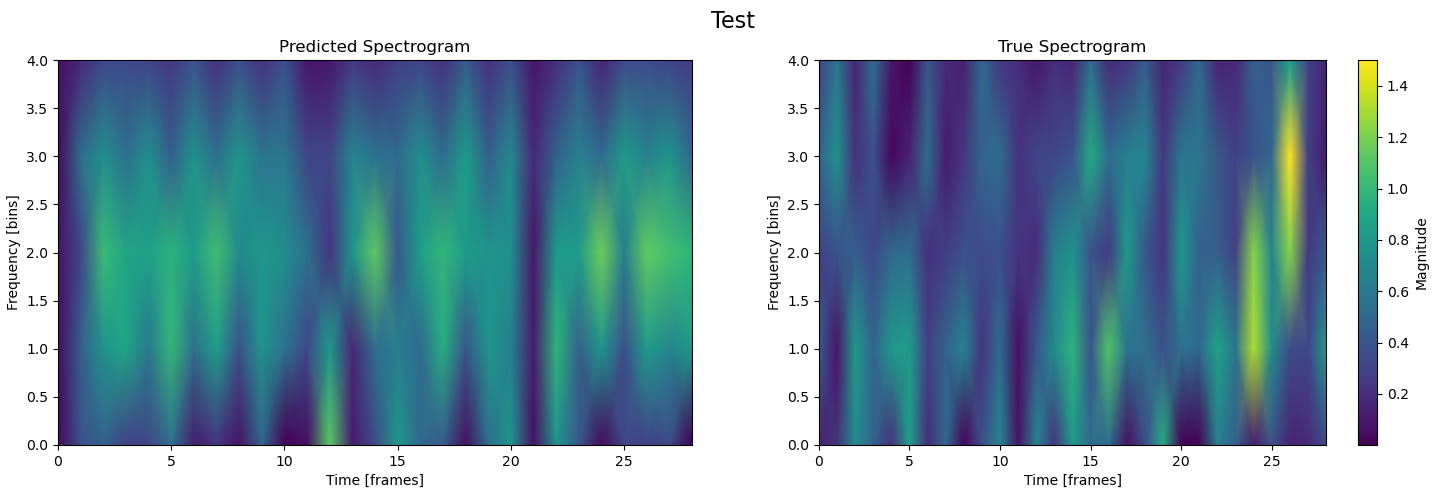}}
    \caption[Predicted spectrogram (test)]{Predicted and true spectrogram. Test results. The predicted spectrogram maintains its accuracy during the test phase (\ac{mse}= 0.55), showcasing the model's generalization ability.}
    \label{fig:spec_test}
\end{figure}

Figures \ref{fig:spec_valid} and \ref{fig:spec_test} illustrate the model's performance in the frequency domain for two individual signals. The predicted spectrograms are compared to the true spectrograms for both validation and test signals.

In both validation (Figure \ref{fig:spec_valid}) and test (Figure \ref{fig:spec_test}) results, the predicted spectrograms closely match the true spectrograms. This demonstrates the model's effectiveness in capturing the frequency characteristics of the signals, a critical aspect for accurate signal analysis in railway maintenance applications.

\begin{figure}[H] \centering
    {\includegraphics[width=\columnwidth]{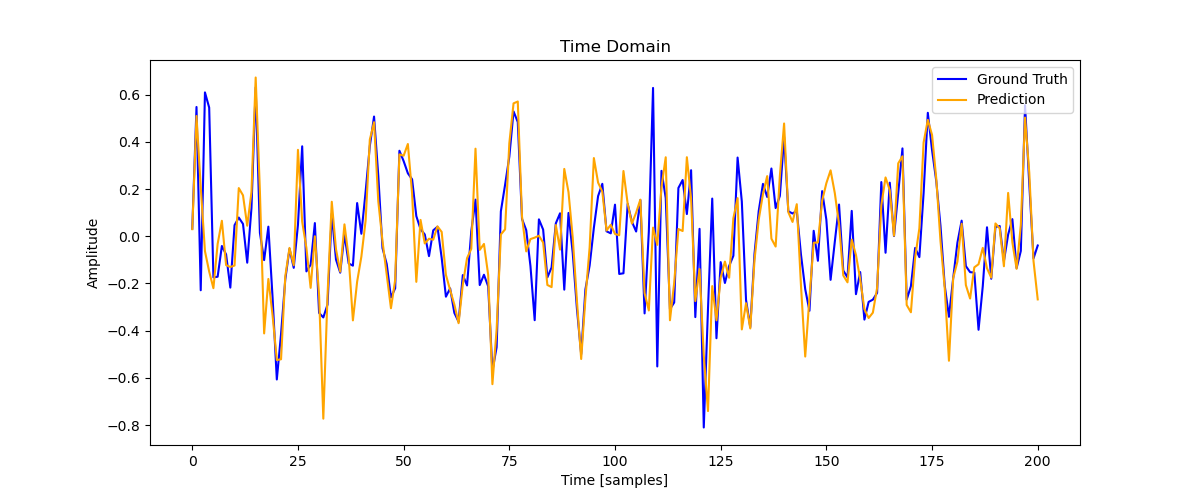}}
    \caption[Predicted time signal (validation)]{Predicted and true signal in the time domain. Validation results. The model's predictions (in orange) align well with the true signal (in blue), demonstrating its effectiveness in time domain forecasting during validation. \ac{mse} = 0.18}
    \label{fig:timepred_nope_val}
\end{figure}

\begin{figure}[H] \centering
    {\includegraphics[width=\columnwidth]{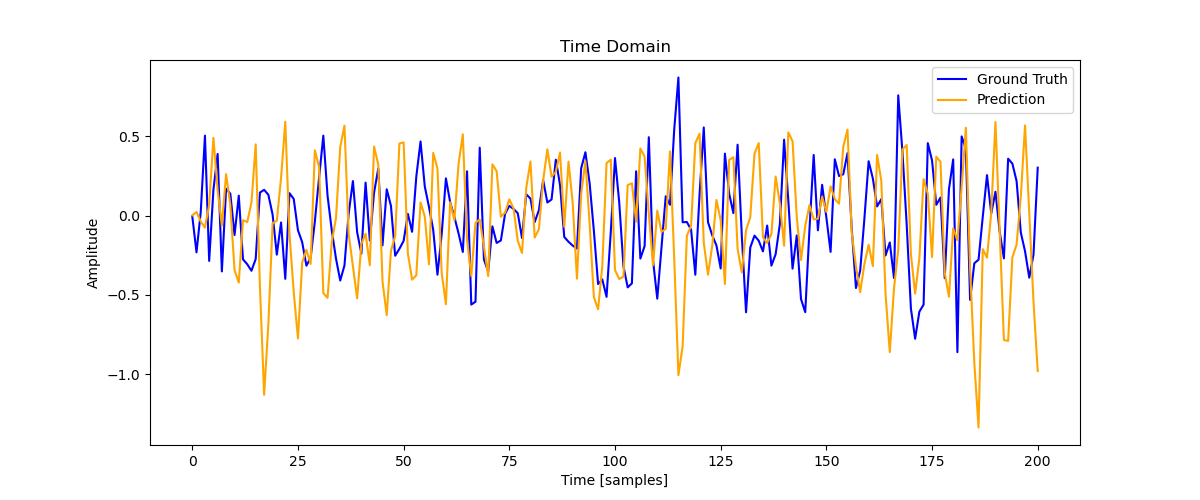}}
    \caption[Predicted time signal (test)]{Predicted and true signal in the time domain. Test results. The alignment of the predicted signal (in orange) with the true signal (in blue) indicates the model's robust performance in time domain forecasting during testing. \ac{mse} = 0.55}
    \label{fig:test_time}
\end{figure}

Figures \ref{fig:timepred_nope_val} and \ref{fig:test_time} show the model's performance in the time domain, comparing the predicted and true signals for both validation and test datasets.

The time domain results indicate that the model's predictions are closely aligned with the true signals for both validation (Figure \ref{fig:timepred_nope_val}) and test (Figure \ref{fig:test_time}) datasets. This alignment confirms the model's robustness in forecasting time-domain signals, which is crucial for predicting potential faults and initiating maintenance actions.

\subsection{Signal Decomposition Analysis}

\begin{figure}[H] \centering
    {\includegraphics[width=\columnwidth]{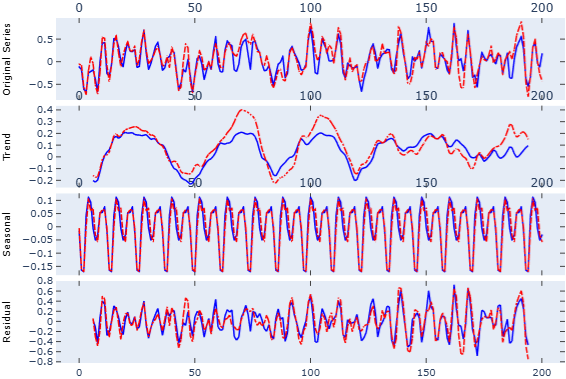}}
    \caption[STL Decomposition (Ground truth)]{Signal STL decomposition. Ground truth shown in blue and predicted signal in red. The signals in the picture correspond to 1) the signal itself 2) the trend component 3) the seasonal component 4) the residual component.}
    \label{fig:decomposition_train}
\end{figure}

To visually analyze the signals, we decompose the signals into their trend, seasonal, and residual components using \ac{LOESS}. Figure \ref{fig:decomposition_train} presents the decomposition for the ground truth and predicted signals, respectively.

The decomposition  shows that the predicted signals retain the essential characteristics of the true signals across all components: trend, seasonal, and residual. This detailed breakdown confirms that the model not only forecasts the overall signal accurately but also captures the underlying patterns effectively.

\subsection{Model Performance Comparison}

Table \ref{table:results_comparison} provides a quantitative comparison of the \ac{sf} and \ac{ssf} models using \ac{mse} as the performance metric.

\begin{table}[H]
    \centering
    {\begin{tabular}{|l|l|l|l|}
    \hline
    \textbf{Model} & \textbf{Train Loss} & \textbf{Validation Loss} & \textbf{Test Loss} \\
    \hline
    \ac{sf} & 0.42811 & 0.43524 & 1.73285 \\
    \ac{ssf} & \textbf{0.1779} & \textbf{0.17223} & \textbf{0.5303} \\
    \hline
    \end{tabular}}
    {\caption{Comparison of models using MSE loss.} \label{table:results_comparison}}
\end{table}

The results in Table \ref{table:results_comparison} demonstrate that the SSF significantly outperforms the \ac{sf} model, achieving lower MSE across training, validation, and test datasets. This improvement underscores the effectiveness of incorporating spectral methods and an improved observation model in enhancing the model's forecasting accuracy and generalization capabilities.

Overall, the figures and table collectively illustrate the superior performance of the SSF model, highlighting its potential for reliable and accurate signal forecasting in railway maintenance applications.

The model addresses missing data imputation by leveraging its capacity to predict missing values based on learned temporal and spectral patterns. During training, the model is exposed to complete datasets with artificially masked data to learn how to reconstruct incomplete sequences. When applied to real-world data, the encoder processes the available portions of the time series to generate a contextual representation, while the decoder predicts the missing values. For extended gaps in data, the model can iteratively predict intermediate points to refine accuracy. By operating in the frequency domain, the model ensures that imputed values preserve the spectral characteristics of the original signal, maintaining consistency in the reconstructed data.

For outlier detection, the model uses its learned patterns to identify deviations between predicted and observed values. The encoder-decoder structure generates expected values based on historical data, and discrepancies between these predictions and actual measurements are flagged as potential outliers. This is possible due to the probabilistic nature of the generative model. The frequency-domain analysis allows the model to detect anomalies in spectral patterns, such as unexpected peaks or harmonic changes, which may indicate system irregularities.

\section{Conclusion}
This study introduces an innovative model architecture designed to enhance the analysis and processing of time series data in the railway industry. By leveraging cutting-edge technologies such as deep learning and the Transformer framework, the proposed model demonstrates significant improvements in performance, efficiency, and adaptability. The incorporation of frequency-domain transformations and the splitting of the encoder into high- and low-frequency blocks results in a highly expressive and efficient representation of input signals. Beyond predictive maintenance, the model offers practical applications, including outlier detection and missing data imputation, showcasing its versatility in addressing key challenges faced by railway monitoring systems. These advancements lay the groundwork for further innovation, such as the development of a bogie digital twin, ultimately contributing to the safety, reliability, and efficiency of modern railway operations.

\section{Discussion}
The proposed model architecture builds upon and extends recent advances in Transformer-based deep learning methods, addressing domain-specific challenges in time series analysis for railway systems. By adapting ideas from the spectral domain and incorporating a novel frequency-based encoder structure, the model achieves a balance between computational efficiency and predictive accuracy. The decision to operate in the frequency domain enables the model to better capture the underlying patterns in the data, aligning with recent research trends.

The applications of this work extend far beyond traditional maintenance. The ability to detect outliers positions the model as a proactive tool for anomaly detection, enabling real-time responses to potential system failures. Furthermore, the model's capacity for missing data imputation addresses a critical limitation of many monitoring systems, ensuring continuous and accurate insights even in the presence of sensor or communication issues. These features enhance the reliability of railway operations and present opportunities for integration into broader digital twin frameworks, potentially transforming maintenance and operational strategies.

Future work could explore optimizing the model for deployment in real-world systems, including hardware constraints and latency requirements. Additionally, expanding the architecture to incorporate multimodal data, such as environmental conditions and operational parameters, could further enhance its applicability and robustness.

\section{Acknowledgements}

This work has been developed within the project Sistema de monitorización de estado para detección de fisuras en ejes ferroviarios (SMEPDFEF-CM-UC3M), 2021 Call for grants to carry out interdisciplinary R\&D projects for young doctors of the Carlos III University from Madrid. The work was also supported by MCIN/AEI/10.13039/501100011033/FEDER,
UE, under grant PID2021-123182OB-I00 and by the Comunidad de Madrid [ELLIS Madrid Unit and IND2022/TIC-23550];

\bibliographystyle{elsarticle-harv.bst}
\bibliography{cas-refs.bib}

\appendix

\label{appendix}

\section{Differentiable sampling operations} 

\subsection{Reparameterization trick} \label{appendix:reparam_trick}

The reparameterization trick is a popular tool used in deep generative models such as VAEs. It is a simple yet powerful trick that allows sampling from the normal distribution while maintaining the gradients propagation.

Instead of sampling from a normal distribution with mean $\mu$ and variance $\sigma^2$, sample from the $N(0,1)$, shift, and scale the sample. This way,  the stochastic process is gathered within a leaf node of the computational graph (an end node), while the random sample is still transmitted to the path within the graph where gradients can backpropagate.

\begin{align}
    X \sim N(\mu, \sigma^2) &\longrightarrow  \text{Sample: x} \\
    Y \sim N(0, 1) &\longrightarrow  \text{Sample: y} \longrightarrow X = \mu + \sigma^2\cdot y
\end{align}

\begin{figure}[H] \centering
    {\includegraphics{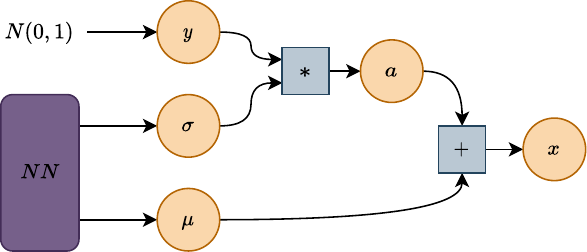}}
    \caption[reparameterization trick]{reparameterization trick. Gradients can propagate from the outputs to the inputs. The random sampling process is performed in a secondary branch of the computational graph.}
\end{figure}

\subsection{Inverse transform sampling}\label{appendix:inverse_transform}

Inverse transform sampling generates samples of a distribution given its cumulative distribution function. 

Consider the exponential distribution with cumulative distribution function  $$F_X(x) = 1 - e^{-\lambda x}$$ for \( x \geq 0 \) (and 0 otherwise). Solving \( y = F(x) \) yields the inverse function  $$x = F^{-1}(y) = -\frac{1}{\lambda} \ln(1-y)$$ This implies that if we generate a random sample \( y_0 \) from a uniform distribution \( U \sim \mathrm{Unif}(0,1) \), and compute $$ x_0 = F_X^{-1}(y_0) = -\frac{1}{\lambda} \ln(1-y_0) $$ then \( x_0 \) follows an exponential distribution.

\begin{figure}[H] \centering
    \includegraphics{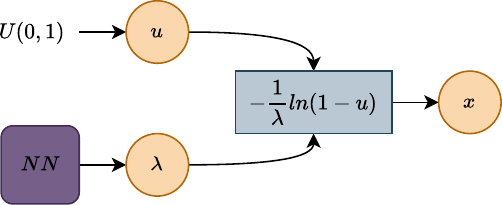}
    \caption[Inverse transform sampling for the exponential distribution: gradient flow visualization]{Inverse transform sampling for the exponential distribution: gradient flow visualization. The gradients flow uninterruptedly from the outputs to the inputs.}
    \label{fig:its_grad}
\end{figure}

As a result, the sampling process can be performed without concerns about the gradient propagation. Summarizing the reparameterization trick computational graph results in Figure \ref{fig:its_grad}. Clearly there exists a differentiable path connecting the inputs and outputs.

\section{Positional encodings}

To add information about the order of the elements,  \citet{vaswani2017attention} use sine and cosine functions of different frequencies to calculate the Positional Encodings (PE):

\begin{equation}
    PE_{(pos, 2i}) = sin(pos/10000^{2i/dmodel})
\end{equation}

\begin{equation}
    PE_{(pos, 2i+1}) = cos(pos/10000^{2i/dmodel})
\end{equation}
where \textit{pos} is the position and \textit{i} is the dimension.

\begin{figure}[H] \centering
    {\includegraphics[width=\columnwidth]{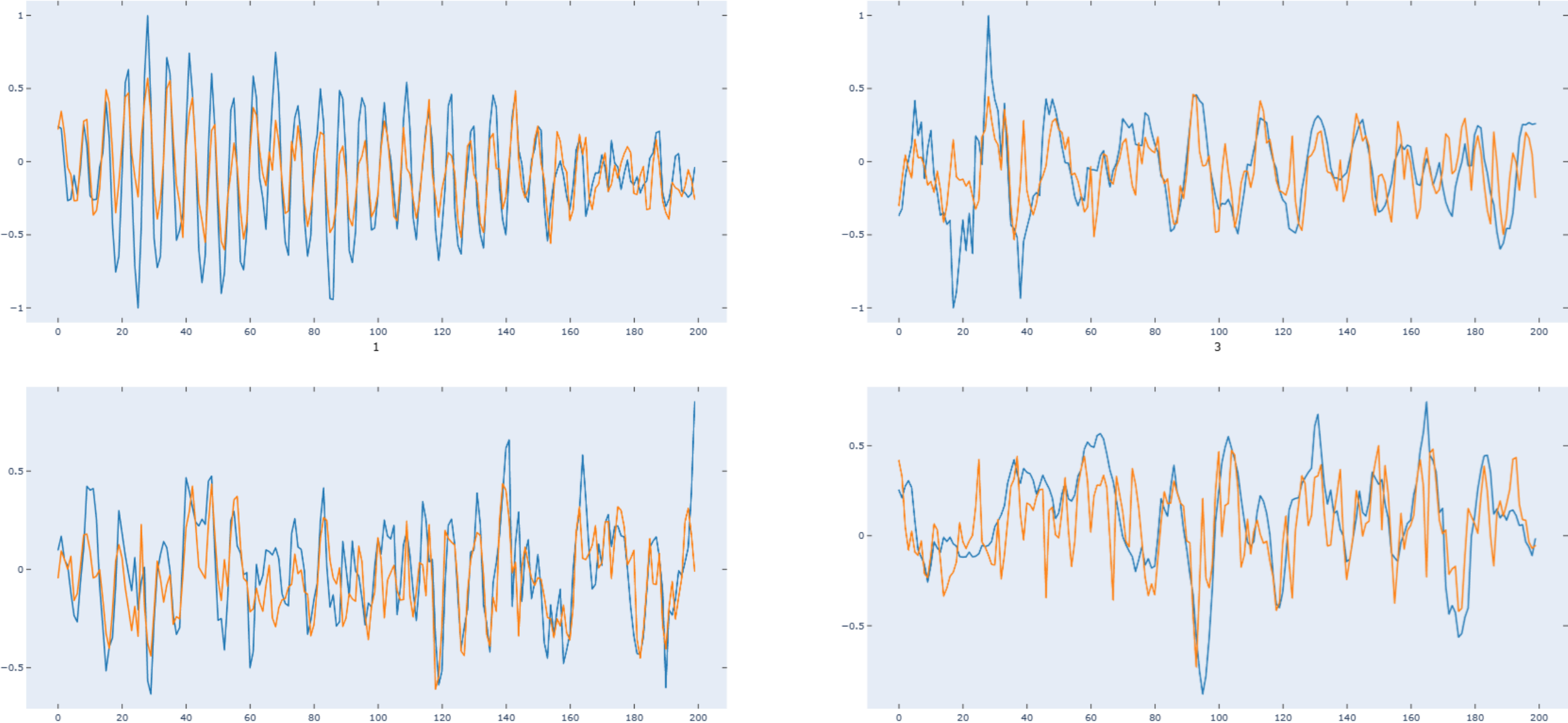}}
    \caption[Result comparison with and without Positional Encoding]{Result comparison with and without Positional Encoding for four different signals. In this figure, no positional encoding has been used.}
\end{figure}

\begin{figure}[H] \centering
    {\includegraphics[width=\columnwidth]{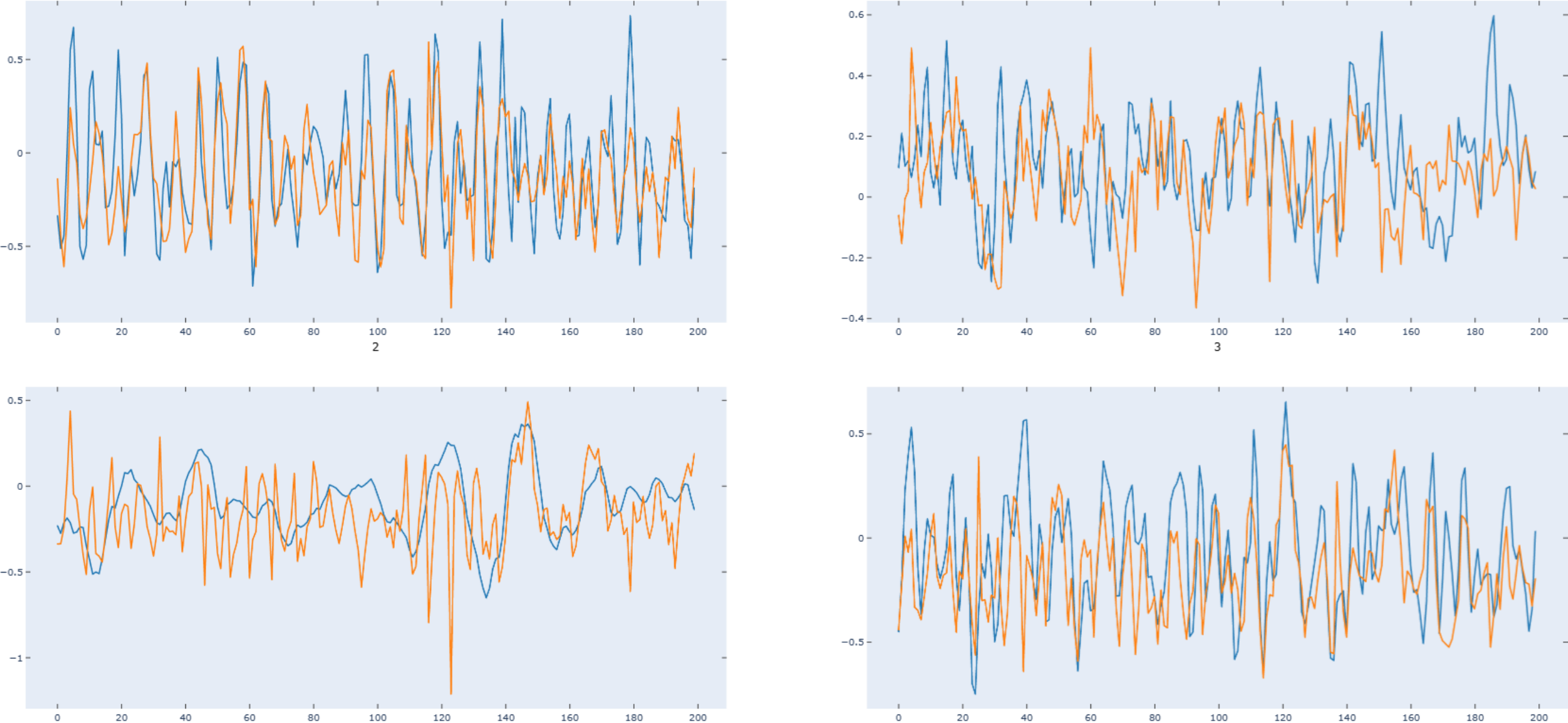}}
    \caption[Result comparison with and without Positional Encoding]{Result comparison with and without Positional Encoding for four different signals. In this figure, the sine and cosine absolute positional encoding has been used.}
\end{figure}

The positional encoding is used as presented above and added to the spectrogram. However, we found that even when removing the positional information, the model still performs well and inherently learns it. This observation is in line with the work presented in \citet{Haviv2022TransformerLM}.

\section{Training} \label{appendix:training}

The training phase was performed using four NVIDIA RTX 4090 GPUs, and intensive hyperparameter cross-validation was performed by Bayesian Optimization method. This method uses a \ac{tpe} to model the relationship between the parameters and the validation loss and chooses the next parameters based on this relationship. Table \ref{table:hparams} summarizes the tested domains for each hyperparameter, explored using \ac{tpe}.

\begin{table}[H] \centering
    {\begin{tabular}{|l|l|l|l|}
    \hline
    \textbf{Hyperparameter} & \textbf{Min} & \textbf{Max} & \textbf{Best Value} \\
    \hline
    Average Pooling Kernel Size & 3 & 27 & 15 \\
    Average Pooling Stride & 1 & 5 & 2 \\
    \textquote{Hi} Attention Dilation & 0 & 15 & 5 \\
    \textquote{Lo} Attention Dilation & 15 & 25 & 14 \\
    Dropout & 0.1 & 0.6 & 0.506 \\
    Freq. Indep. 1-D Conv. Kernel Size & 3 & 27 & 15 \\
    Number of Heads ('Hi' Attention) & 1 & 6 & 1 \\
    Number of Heads ('Lo' Attention) & 1 & 6 & 5 \\
    Time Resolution (\ac{stft}) & 8 & 16 & 7 \\
    Learning Rate & 1e-2 & 1e-4 & 2.97881e-3 \\
    Mean Kernel Size & 3 & 27 & 17 \\
    Number of Encoder Layers & 1 & 4 & 3 \\
    Frequency Resolution & 1 & 30 & 2 \\
    Time Compression & 1 & 5 & 3 \\
    Variance Kernel Size & 3 & 27 & 15 \\
    \hline
    \end{tabular}}
    {\caption{Tested Hyperparameters Range and Best Values.} \label{table:hparams}}
\end{table}

For each of the datasets WS1, WS2 and WS3, the model was trained using 70\% of the data, validated using 20\%, and tested using the last 10\%. The model performs similarly regardless of the used dataset. 

The models were trained for a maximum of 1000 epochs.

\begin{figure}[H] \centering
    \includegraphics[width=\columnwidth]{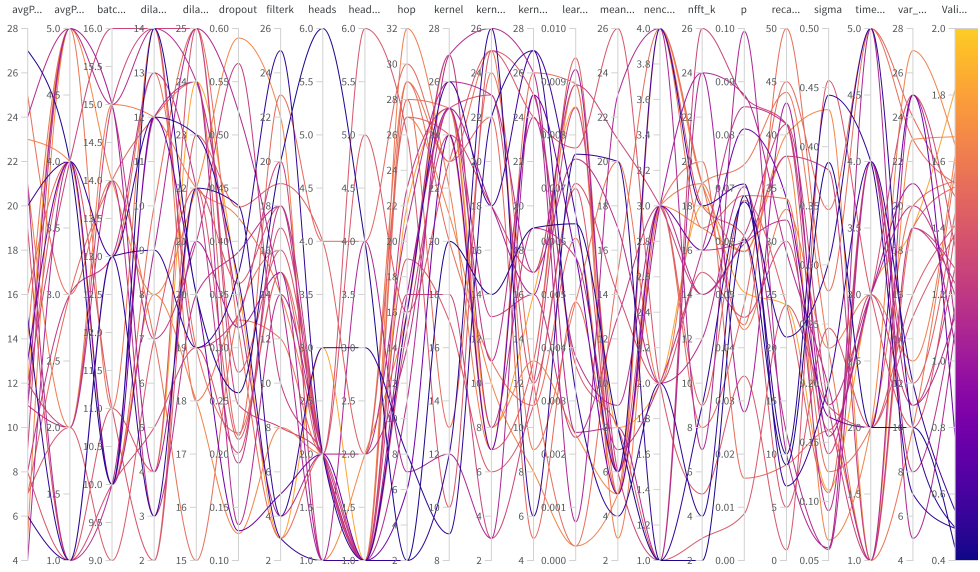}
    \caption[Sweep plot]{Sweep plot with some of the combinations of hyperparameters tested. The vertical axes contain the domain of the hyperparameters. Each line crosses these axes through the values that characterized each run. The rightmost axis represents the obtained validation loss.}
    \label{fig:sweep}
\end{figure}

Figure \ref{fig:sweep} provides a diagram containing some of the combinations and their resulting validation losses.

\end{document}